\documentclass[12pt]{amsart}

\usepackage{amsfonts, amssymb, amsthm, amsmath, bbm, wasysym, enumerate, amsxtra, latexsym, fullpage, amscd, graphics, epic, sansmath, mathtools, multicol, url, hyperref, bbm, blindtext, wasysym, comment, datetime, stmaryrd}
\usepackage[shortlabels]{enumitem}
\usepackage[all,cmtip]{xy}
\usepackage[dvipsnames]{xcolor}
\usepackage[makeroom]{cancel}  

\usepackage[hang,small,bf]{caption}    

\usepackage{tikz}	
\usetikzlibrary{backgrounds,fit,decorations.pathreplacing}  

\theoremstyle{plain}
\newtheorem{theorem}{Theorem}[section]

\theoremstyle{remark}

\newtheorem{example}[theorem]{Example}

\newcommand{\person}{\mathop{\texttt{person}}\nolimits}
\newcommand{\child}{\mathop{\texttt{child}}\nolimits}
\newcommand{\constraint}{\mathop{\textrm{constraint}}\nolimits}

\def\BibTeX{{\rm B\kern-.05em{\sc i\kern-.025em b}\kern-.08em
    T\kern-.1667em\lower.7ex\hbox{E}\kern-.125emX}}
    
\title{Genetic optimization algorithms applied toward mission computability models}
\author[Mee Seong Im]{Mee Seong Im}
\address{Department of Mathematical Sciences, U.S. Military Academy, West Point, NY 10996} 
\email{meeseong.im@westpoint.edu, meeseongim@gmail.com}
\author[Venkat R. Dasari]{Venkat R. Dasari}
\address{U.S. Army Research Laboratory, Aberdeen Proving Ground, MD 21005}
\email{venkateswara.r.dasari.civ@mail.mil}

\begin{document}

\begin{abstract}
Genetic algorithms are modeled after the biological evolutionary processes that use natural selection to select the best species to survive. They are heuristics based and low cost to compute. Genetic algorithms use selection, crossover, and mutation to obtain a feasible solution to computational problems.  In this paper, we describe  our  genetic optimization algorithms to a mission-critical and constraints-aware computation problem. 
\end{abstract}

\maketitle

\section{Introduction} 
\label{section:intro}
Mission critical computations are often needed in the tactical edge  where the computational and  energy resources are limited and impose a limit on the computational complexity of a chosen algorithm to accomplish mission computational objectives.  A trade off must be made between the degree of optimization, speed of the computation and the minimal computational accuracy needed to achieve the computation objectives like decision making, objection detection and obstacle detection. Genetic algorithms are well suited to deploy in resource constrained tactical environments  because they are fast, use  low resources and their degree of computation accuracy is  just enough to meet mission computation objectives in many cases.

In \cite{Im-Dasari-quantum-communication-channels}, Im and Dasari analyze and optimize the parameters of the communication channels needed for quantum applications to successfully operate via classical communication channels. We also develop algorithms for synchronizing data delivery on both classical and quantum channels. 
The authors in \cite{dasari2018complexity}, together with Billy Geerhart, investigate computational efﬁciency in terms of watt and computational speeds to match mission requirements in mission-oriented tactical environments since it plays an important role in determining the ﬁtness of an application for mission deployment; the application we study in this manuscript relates to context-aware and mission-focused computational framework by restricting to a subclass of deterministic polynomial time complexity class of languages. We focus on a constraints-aware distributed computing algorithm, where if too many jobs are assigned to an array of cores, an algorithm can be written to minimize the number of failed jobs.

The authors address the
complexity of various optimization strategies related to low SWaP computing in \cite{IDBS-SWAP-tactical-computing}, and due to the restrictions toward  optimization of computational resources and intelligent time versus eﬃciency tradeoﬀs in decision making, only a small subset of less complicated and fast computable algorithms can be used for tactical, adaptive computing.  And an eﬃciency tradeoﬀ we consider in this manuscript is a job scheduling system for batch jobs and remote desktops for graphical applications, and discuss a way to optimize the heterogeneous computing environment. 
In  
\cite{DIB-quantum-comp}, Dasari--Im--Beshaj discuss classical image processing algorithms that will benefit from quantum parallelism. 

Genetic algorithms mimic ideas from evolutionary biology techniques (cf.~\cite{whitley1994genetic,maulik2000genetic,harik1999compact,deaven1995molecular,yang1998feature,goldberg1986engineering,ribeiro1994genetic}).  
An instance of such algorithms is called ant colony optimization technique (cf.~\cite{dorigo2006ant,dorigo1999ant,dorigo2005ant,blum2005ant}). 
In this manuscript, we investigate a mission computability model that will benefit from genetic optimization algorithms.

\section{Bio-inspired algorithms} 
\label{section:bio-inspired-algor}
Genetic algorithms begin with an initial generation of feasible solutions that are tested against the objective function. Subsequent generations evolve through notions called selection, crossover, and mutation.   
 Selection is to retain the best performing (binary) bit strings from one generation to the next (see e.g., Example~\ref{ex:persons-child-genetic-algorithm}). 
Crossover is a notion where we select common similarity between the different $\person$ variables and keep those to be the same to create a $\child$ variable that appear in the subsequent generation. 
Finally, mutation is where we mutate certain variables from $\person$ to take on random values and create a $\child$ based off of the mutation. Mutation allows genetic algorithms to avoid falling into a local minima and helps them to explore the space of possible solutions. 

\begin{example}
\label{ex:persons-child-genetic-algorithm}
Consider 
\begin{align*}
\person1 &= 
\begin{pmatrix}
0 & 0 & 1&0 &1 &0 &1 &0 & 1& 0  
\end{pmatrix} \quad \mbox{ and } \\ 
\person2 &= 
\begin{pmatrix}
1 & 0 & 1&0 &0 &0 &0 &0 & 1& 1    
\end{pmatrix} 
\end{align*}
from the previous generation, where the binary digits are the variables in the optimization problem. We use selection 
\begin{align*}
\person1 &= 
\begin{pmatrix}
0 & {\color{red}0} & {\color{ForestGreen}1}&{\color{red}0} &1 
&{\color{red}0} &1 &{\color{red}0} & {\color{ForestGreen}1}& 0  
\end{pmatrix} \mbox{ and }\\ 
\person2 &= 
\begin{pmatrix}
1 & {\color{red}0} & {\color{ForestGreen}1}&{\color{red}0} &0 
&{\color{red}0} &0 &{\color{red}0} & {\color{ForestGreen}1}& 1   
\end{pmatrix},  
\end{align*}
and favor them for reproduction; 
thus, red and green digits are kept for the next generation since they performed well in the previous generation. Because they performed well, they might be used for crossover, i.e., 
\begin{align*}
\person1 &= 
\begin{pmatrix}
0 & {\color{red}0} & {\color{ForestGreen}1}&{\color{red}0} &1 
&{\color{red}0} &1 &{\color{red}0} & {\color{ForestGreen}1}& 0  
\end{pmatrix}, \\ 
\person2 &= 
\begin{pmatrix}
1 & {\color{red}0} & {\color{ForestGreen}1}&{\color{red}0} &0 
&{\color{red}0} &0 &{\color{red}0} & {\color{ForestGreen}1}& 1   
\end{pmatrix}, \mbox{ and }  \\ 
\child &= 
\begin{pmatrix}
0 & {\color{red}0} & {\color{ForestGreen}1}&{\color{red}0} &1  
&{\color{red}0} &0 &{\color{red}0} & {\color{ForestGreen}1}& 0   
\end{pmatrix}.  
\end{align*}
Mutations may occur, assisting the genetic algorithm to better analyze the solution set: 
\begin{align*}
\person1 &= 
\begin{pmatrix}
0 & 0 & {\color{ForestGreen}1}&0 & {\color{ForestGreen}1} 
&0 & {\color{ForestGreen}1} & 0 & {\color{ForestGreen}1}& 0  
\end{pmatrix} \mbox{ and } \\ 
\child &= 
\begin{pmatrix}
{\color{ForestGreen}1} & {\color{ForestGreen}1} & 0 & 0 &{\color{ForestGreen}1}  
& {\color{ForestGreen}1} &0 &0 & {\color{ForestGreen}1}& {\color{ForestGreen}1}  
\end{pmatrix}.  
\end{align*}

\end{example}
 
\section{Optimization of genetic algorithms} 
\label{section:genetic-optimization} 
Optimization of genetic algorithms can be thought of as follows.  
Consider an optimization problem in an $n$-dimensional space,  
with a given initial population $s_1^{(1)},\ldots, s_N^{(1)}$, i.e., the first generation solution set, scattered in the $n$-dimensional space. Suppose there are $k$ local minima in the initial population, where, without loss of generality, all are local $m_1,\ldots, m_{k-1}$ and the last one is global, which we denote by $m_k$.

The fitness function $f$ evaluates the first generation solution set, giving us fitness function values $f(s_1^{(1)}),\ldots , f(s_N^{(1)})$. A subset of the values are selected as a few good solutions (for e.g., one can use the distance function and a few good solutions are considered to be those solutions $s_i^{(1)}$ satisfying the inequality $d(f(s_i^{(1)}),m_j)=|\!|f(s_i^{(1)}) - m_j |\!|\leq \varepsilon$ where  $1\leq i\leq N$ and $1\leq j\leq k$, and $\varepsilon>0$). Now using the good solutions, we use selection, crossover, and mutation to generate a new population $s_1^{(2)},\ldots, s_N^{(2)}$ of feasible solutions. They are then evaluated back into the fitness function, and we repeat this process of generating new generations of solutions until the genetic algorithm converges through a multitude of convergence criteria. 

Some examples of convergence criteria include fixing the number of generations so that the genetic algorithm will run until we compute a certain number of generations. 
Another one is the genetic algorithm will converge and will come to a halt when the \textit{best} objective function or \textit{best} fitness function value is no longer changing or it is altering infinitesimally.

\section{Mission computability models}
\label{section:mission-computability-models}

Mission computability has been explored in \cite{Im-Dasari-quantum-communication-channels,dasari2018complexity,IDBS-SWAP-tactical-computing,DIB-quantum-comp,mackenzie1997multiagent,discrete-event-command,brucker2012emerging}, to name a few. 
Fulfilling computational requirement for a given mission is crucial for a successful Army operation. Algorithms generally studied in terms of time and space complexity using limited number of resources often fail due to inefficiency and constraints of the computation or resource. 
Network specific constraints also need to be taken account when assessing resource efficiency of the distributed computation.

\section{Applications}
\label{section:applications}

In this manuscript, we consider a genetic optimization algorithm for the class of mission computable problems (cf.~\cite{dasari2018complexity,IDBS-SWAP-tactical-computing}). That is, consider a constraints-aware distributed computation: 
given that many jobs are assigned to an array of cores, we incorporate an algorithm to minimize the number of failed jobs.

We program $k$ machines, each with a single core, using integer programming. That is, we arrange the decision variables in an ordered list which is represented as a binary matrix $M= (b_{i,j})_{i,j}\in M_{m+1,k+1}$, where $b_{i,j}\in \mathbb{Z}_2$, since we want to maximize the total number of computations on $k$ machines, where $(i,j)$ represents the $i$-th job on machine $j$ for $i\geq 2$. 
We reserve the first column of $M$ to label and distinguish the job, and let us reserve the first row of $M$ to be those computations that have successfully been computed in machine $j$. 
For the sake of notational simplicity, we label the columns of $M$ by $0$ through $k$ (so the $j$-th column corresponds to $j$-th machine).
A $1$ in $(i,j)$ position in matrix $M$ represents that the $i$-th job has been sent to machine $j$ and is waiting to be processed, while a $0$ in $(i,j)$ position represents that the $i$-th job has failed to be sent to machine $j$. 
Once the computations in row $1$ are complete, all $b_{i,j}$ should move up to position $b_{i-1,j}$ for all $i$ and $j$. This action could be viewed as a left multiplication by the $m\times m$ nilpotent matrix 
\[ 
u = 
\begin{pmatrix} 
0 & 1 & \ldots & 0 \\ 
\vdots & \ddots & \ddots & \vdots \\ 
0 & \ldots & \ldots & 0 \\ 
\end{pmatrix} 
\]  
onto the $m\times k$ lower right submatrix of $M$.

Since we want the $i$-th job to successfully wait in-line by at most one machine,  we have a constraint equation 
\[ 
\constraint\!: \qquad \sum_{j=1}^k b_{i,j} \leq 1. 
\]  
We have another constraint condition since we want each job to be successfully computed at most once 
\begin{center}
$\constraint\!:$ \\ 
if $b_{1,\ell}=1$ for some $\ell$, \\ 
then remove this job from being called again, \\ 
but if $b_{1,\ell}=0$ for all $\ell$, then recall this job \\ 
by putting it in any entry in the last row of matrix $M$ \\ 
(so the coordinate being $b_{m+1,j}=1$ for exactly one $j$).
\end{center}
 
Now we define a notion of inverse selection.  If machines $\ell_1,\ldots, \ell_{u}$ have more than $w$ jobs waiting, i.e., $\sum_{i=2}^{m+1} b_{i,j}>w$ for $j=\ell_1,\ldots, \ell_u$ (that is, for certain subset of the columns), then we use inverse selection to identify these columns and then use crossover to create a new matrix $M'$ to minimize the number of columns satisfying $\sum_{i=2}^{m+1} b_{i,j}>w$ such that $u\leq \delta$ for some predetermined $\delta > 0$. This eases some of the machines from having too many jobs lined up and from overworking, and possibly breaking down or needing maintenance. 
Furthermore, mutation is applied to those rows of $M$ with all $0$ such that exactly one $1$ appears in such rows, so that dropped jobs are minimized.

Another application is using ant colony optimization algorithm to mission computability models. Ant colony optimization algorithm uses crowding technique and changeable mutations to multiple populations to obtain near-optimal solutions (cf. \cite{chong2010continuous,davis2017approximate,wang2007modeling,fu2006improved,lee2002immunity,zhang2005ant}). It can be thought of as an existence of a higher probability that a population will ultimately \textit{move along a path more frequently traveled}. The ant colony algorithm can be run continuously, mutate, and adapt in real time, even under incremental modifications and configurations of the system.  Some examples include taking the shortest path from point $A$ to point $B$, even if points $A$ and $B$ were to move around in real time, or photons traveling in polarization-maintaining cables for communication purposes.

\section{Conclusion} 
\label{section:conclusion} 
 
We have discussed genetic algorithms and mission computability, applying genetic optimization techniques to an important computability and processing problem. In particular, genetic algorithm has been applied to jobs sent to and scheduled on a distributed computing platform, where a global optimizer with constraints determines which jobs should be scheduled or rescheduled. The notion of crossover has been used to move the jobs \textit{up} along an integral matrix, and since each machine tags each job with appropriate priority level, the notion of selection and mutation in genetic algorithms has been applied to handle dropped jobs.

Applying genetic and ant colony optimization algorithms to many crucial mission-critical problems is a quickly developing field, which we foresee that these techniques will become an important contribution and development in this area of research.

\section*{Acknowledgements}
\label{sectiion:ackn}

This work is supported by a research collaboration between the U.S. Army Research Laboratory and U.S. Military Academy. M.S.I. is supported by National Research Council Research Associateship Programs.

\bibliographystyle{IEEEtran}
\bibliography{qml}

\end{document}